\documentclass[journal]{IEEEtran}
\usepackage{amsmath}
\usepackage{algorithmicx,algorithm}
\usepackage[noend]{algpseudocode}
\usepackage{subfigure}
\usepackage{booktabs}
\usepackage{multirow}
\usepackage{amssymb}
\usepackage{hyperref}

\usepackage{float}

\usepackage{color,xcolor}
\definecolor{reds}{RGB}{184,84,80}
\definecolor{blues}{RGB}{16,115,158}
\usepackage{graphicx}
\usepackage{caption} 

\definecolor{highlight}{rgb}{1,1,1} 
\definecolor{high}{rgb}{0,0,0} 

\ifCLASSINFOpdf
\else
\fi

\usepackage{times}
\usepackage{latexsym}

\usepackage[T1]{fontenc}

\usepackage[utf8]{inputenc}
\usepackage{microtype}

\usepackage{times}
\usepackage{latexsym}
\usepackage{multicol}
\usepackage{subfigure}
\usepackage{extarrows}
\usepackage{algorithm}  
\usepackage{algorithmicx}  
\usepackage{algpseudocode}  
\usepackage{amsmath} 
\usepackage{color,xcolor}
\usepackage{graphicx}
\usepackage{booktabs}

\definecolor{red}{RGB}{184,84,80}
\definecolor{blue}{RGB}{16,115,158}

\usepackage{scalerel}
\usepackage{tikz}
\usetikzlibrary{svg.path}

\definecolor{orcidlogocol}{HTML}{A6CE39}
\tikzset{
    orcidlogo/.pic={
        \fill[orcidlogocol] svg{M256,128c0,70.7-57.3,128-128,128C57.3,256,0,198.7,0,128C0,57.3,57.3,0,128,0C198.7,0,256,57.3,256,128z};
        \fill[white] svg{M86.3,186.2H70.9V79.1h15.4v48.4V186.2z}
        svg{M108.9,79.1h41.6c39.6,0,57,28.3,57,53.6c0,27.5-21.5,53.6-56.8,53.6h-41.8V79.1z M124.3,172.4h24.5c34.9,0,42.9-26.5,42.9-39.7c0-21.5-13.7-39.7-43.7-39.7h-23.7V172.4z}
        svg{M88.7,56.8c0,5.5-4.5,10.1-10.1,10.1c-5.6,0-10.1-4.6-10.1-10.1c0-5.6,4.5-10.1,10.1-10.1C84.2,46.7,88.7,51.3,88.7,56.8z};
    }
}

\newcommand\orcidicon[1]{\href{https://orcid.org/#1}{\mbox{\scalerel*{
                \begin{tikzpicture}[yscale=-1,transform shape]
                \pic{orcidlogo};
                \end{tikzpicture}
            }{|}}}}

\definecolor{highlight}{rgb}{0,0,0} 

\begin{document}
%
\title{LOGEN: Few-shot Logical Knowledge-Conditioned Text Generation with Self-training}
%
%
%

\author{Shumin Deng$^{\textsuperscript{\orcidicon{0000-0002-4049-8478}}}$, 
        Jiacheng Yang$^{\textsuperscript{\orcidicon{0000-0000-0000-0000}}}$, 
        Hongbin Ye$^{\textsuperscript{\orcidicon{0000-0001-5727-5599}}}$, 
        Chuanqi Tan$^{\textsuperscript{\orcidicon{0000-0002-6676-3057}}}$, 
        Mosha Chen$^{\textsuperscript{\orcidicon{0000-0001-8815-6031}}}$, \\
        Songfang Huang$^{\textsuperscript{\orcidicon{0000-0001-8084-0904}}}$,
        Fei Huang$^{\textsuperscript{\orcidicon{0000-0002-3709-5053}}}$, 
        Huajun Chen$^{\textsuperscript{\orcidicon{0000-0001-5496-7442}}}$
        Ningyu Zhang\dag$^{\textsuperscript{\orcidicon{0000-0002-1970-0678}}}$
\IEEEcompsocitemizethanks{

\IEEEcompsocthanksitem Shumin Deng, Jiacheng Yang, Hongbin Ye are with Zhejiang University, and supported by Alibaba-Zhejiang University Joint Research Institute of Frontier Technologies (AZFT), Hangzhou, China, 310007\protect
\IEEEcompsocthanksitem Shumin Deng is with National University of Singapore, Singapore, 117602\protect
\IEEEcompsocthanksitem Chuanqi Tan, Mosha Chen, Songfang Huang, Fei Huang are with Alibaba Group, Hangzhou, China, 3111121\protect
\IEEEcompsocthanksitem Huajun Chen is with Zhejiang University and Donghai Laboratory, Hangzhou, China, 310007\protect
\IEEEcompsocthanksitem Ningyu Zhang is with Zhejiang University, and supported by Alibaba-Zhejiang University Joint Research Institute of Frontier Technologies (AZFT), Hangzhou, China, 310007\protect\\
E-mail: zhangningyu@zju.edu.cn, corresponding author.\\
}

}


%
%

\markboth{IEEE/ACM TRANSACTIONS ON AUDIO, SPEECH, AND LANGUAGE PROCESSING,~Vol.~14, No.~8, August~2021}{Shell \MakeLowercase{\textit{et al.}}: Bare Demo of IEEEtran.cls for IEEE Journals}

%



\maketitle

\begin{abstract}
Natural language generation from structured data mainly focuses on surface-level descriptions, suffering from uncontrollable content selection and low fidelity. Previous works leverage logical forms to facilitate logical knowledge-conditioned text generation. Though achieving remarkable progress, they are data-hungry, which makes the adoption for real-world applications challenging with limited data. To this end, this paper proposes a unified framework for logical knowledge-conditioned text generation in the few-shot setting. With only a few seeds logical forms (e.g., 20/100 shot), our approach leverages self-training and samples pseudo logical forms based on content and structure consistency. Experimental results demonstrate that our approach can obtain better few-shot performance than baselines.

\end{abstract}


%
\IEEEpeerreviewmaketitle
\section{Introduction}
\begin{figure}[H]
\centering 
\includegraphics[width=0.5\textwidth]{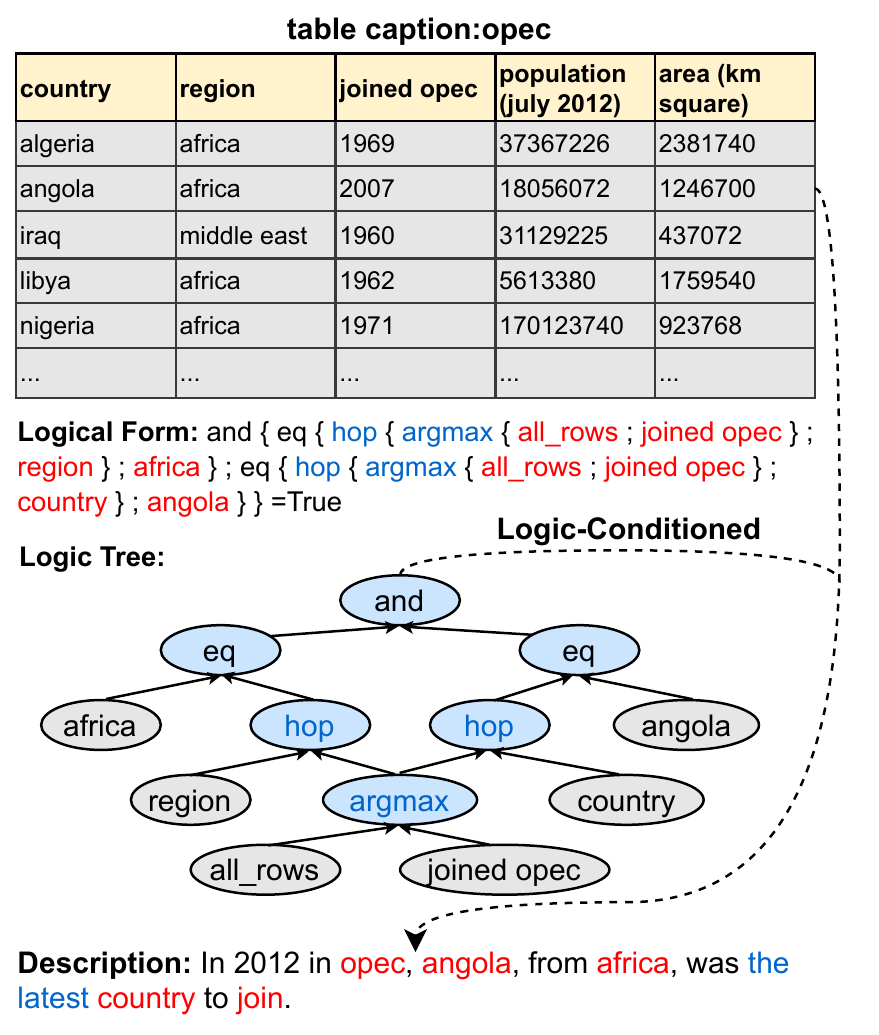} 
\caption{Logical knowledge-conditioned text generation.}
\label{motivation}
\end{figure}

\IEEEPARstart{N}{atural} language generation (NLG) from structured data has good application prospects in communicating with humans in a natural way \cite{DBLP:conf/kdd/ZhangJD0YCTHWHC21}, such as financial report \cite{DBLP:conf/acl/MurakamiWMGYTM17}, medical report \cite{hasan2019clinical} and so on.
However, previous studies \cite{DBLP:conf/coling/GongSFQBLL20} mostly concentrate on surface descriptions from simple records, such as limited schema (e.g., E2E \cite{DBLP:journals/csl/DusekNR20}, and WikiBio \cite{DBLP:conf/emnlp/LebretGA16}), which suffer from low fidelity and uncontrollable content selection. 
Logical forms can condition the generation beyond superficial facts (e.g., ``\emph{Canada has got 3 gold medals}. ’’) with new statements that can be entailed from these facts (e.g.,``\emph{Canada obtained the most gold medals}. ’’). \cite{DBLP:conf/emnlp/ChenCZZZSW20} first leverage logical forms for NLG, and the generation module is provided with the table information and a logical form representing the target text's semantics (see Figure \ref{motivation} for an example).

However, the success of this approach is heavily dependent on the availability of a massive amount of labeled training data, e.g., 10.8k logic-text training pairs for the \textsc{Logic2Text} dataset \cite{DBLP:conf/emnlp/ChenCZZZSW20} in a single domain. Such data-hungry nature makes NLG systems challenging to be widely adopted in real-world applications.  To this end, we focus on exploring how to efficiently model for few-shot logical knowledge-conditioned text generation, which is not well-studied before.

To address the few-shot issue, one of the most potent methods is meta-learning, which transfers the experience learned from similar tasks to the target task \cite{DBLP:conf/icml/FinnAL17,DBLP:conf/naacl/ZhangDSWCZC19,DBLP:conf/www/ZhangDSCZC20}. However, they have difficulty tackling text generation, mainly attributed to the excessive time cost required to train numerous instances. Another intriguing idea is to leverage unlabeled data via semi-supervised learning, which is useful for improving model performance when the target domain lacks manual resources. 
Self-training is a classical, intuitive and straightforward semi-supervised learning method, which first trains the model with labeled data and then enlarges the labeled set according to the most confident predictions (a.k.a., pseudo labels) on unlabeled data \cite{DBLP:conf/nips/OliverORCG18}. 

Note that there exists a quantity of in-domain raw text; it is intuitive to generate pseudo logical forms and leverage those unlabeled data via self-training. However, there are still several nontrivial challenges for self-training in text generation. Since there are only a few parallel data for training, those pseudo logical forms may contain many ill-posed samples, which may no longer bring performance improvement but even deteriorate the performance when noisy instances exceed the model's robustness. Specifically, those pseudo logical forms should guarantee the content and structure consistency. 
{\color{highlight}Content consistency indicates that the generated logical forms have a consistent semantic meaning aligned with the input text. Structure consistency refers to the fact that the symbolic structure conforms to the logical specifications.} 

To alleviate the aforementioned problems, we propose a unified framework for the few-shot \textbf{LO}gical knowledge-conditioned text \textbf{GEN}eration, namely \textbf{LOGEN}. Our approach utilizes self-training to leverage those easy-to-obtain in-domain corpora. Specifically, we utilize a sequence-to-sequence model (e.g., text-to-logic) trained with few-shot seed data to generate pseudo logical forms. To guarantee content consistency and structure consistency, we employ two key components. The \textbf{first} one is a content-consistency module. We utilize the reverse task of pseudo-logical form generation (i.e., target logical knowledge-conditioned text generation) to recover the input text. Our assumption states that, \emph{if the recovered text is semantically similar to the input text, the generated symbolic logical form will be of high quality}. The \textbf{second} component is a structure-consistency module. We design a logical rationality estimator based on the general rules. Specifically, we convert the generated logical forms to trees and leverage domain rules to calculate logical rationality. Finally, we obtain the quality score of each generated logical form. We then select the top-K instances as high-quality logical forms and train the model iteratively until no unlabeled data remain.

In summary, our main contributions include:
\begin{itemize}
\item We study the few-shot logical knowledge-conditioned text generation problem, which is a new branch of research that has not been well-explored to the best of our knowledge.

\item We propose the LOGEN framework, which leverages self-training and samples pseudo-logical forms based on content and structure consistency. 

\item Experimental results on the benchmark dataset illustrate that our approach can achieve better performance than baselines in the few-shot setting.

\end{itemize}
 
\section{Related Work}

\begin{figure*}[h]
\centering

 \includegraphics[width=0.9\textwidth]{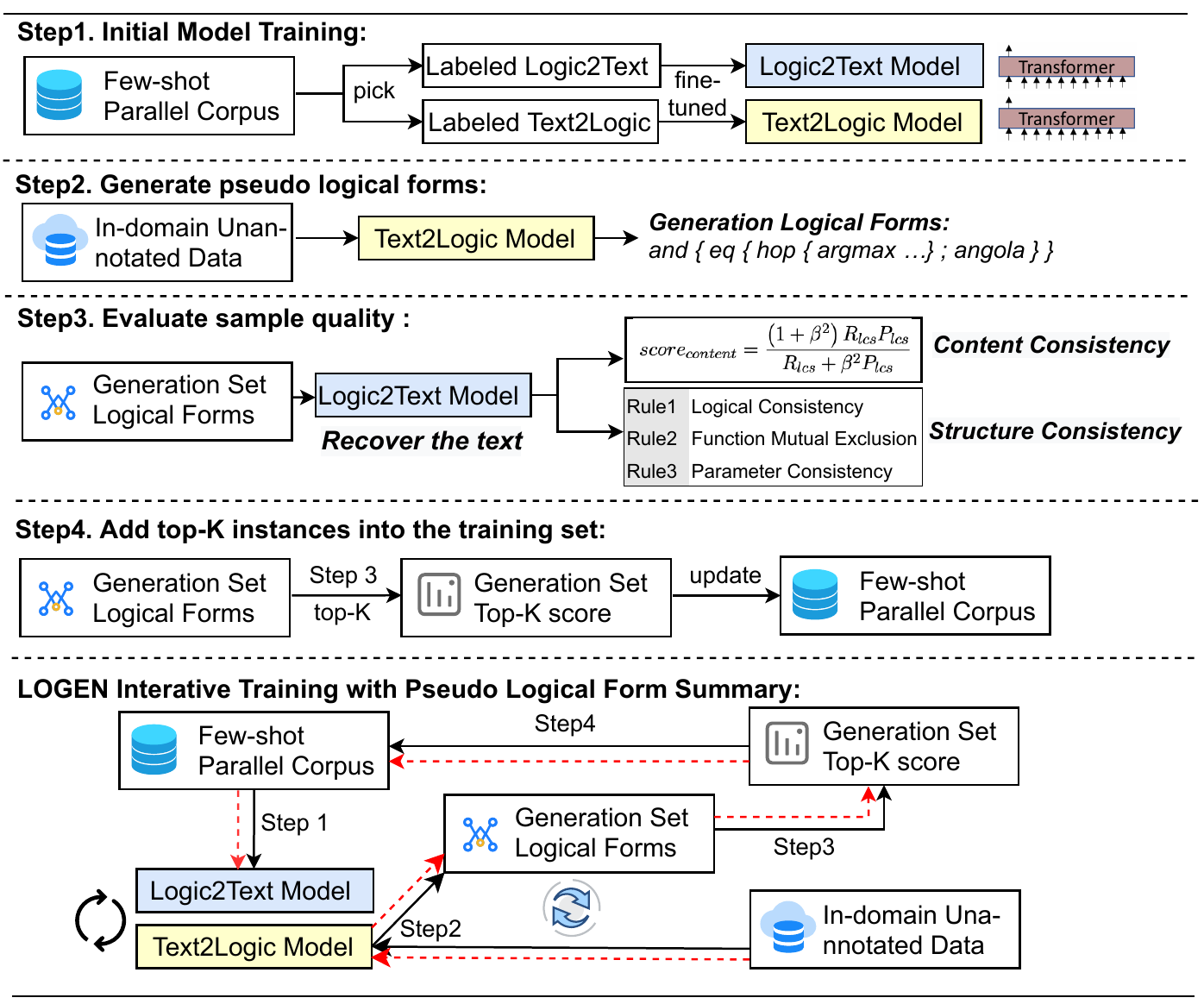}
\caption{The pipeline of the proposed \textbf{LOGEN} framework in the few-shot setting.}
\label{arc}
\end{figure*}

NLG from structured data has been appealed to researchers for many years~\cite{DBLP:conf/acl/ZhaoWC20,DBLP:conf/acl/WangWAYC20,shen-etal-2020-neural,chang-etal-2020-dart,chen-etal-2020-kgpt,shahidi-etal-2020-two}, resulting in many real-world applications,  including those for the automatic generation of weather reports~\cite{DBLP:conf/acl/LiangJK09}, sport reports~\cite{DBLP:conf/emnlp/WisemanSR17}, and clinical reports~\cite{dimarco2007development, lee2018natural}. 
Previous approaches typically utilize pipeline-based approaches that included surface realization and content selection~\cite{DBLP:journals/nle/ReiterD97,DBLP:journals/jair/GattK18}. 
More recent models tend to leverage end-to-end neural network for tasks such as table-to-text generation \cite{DBLP:conf/emnlp/WisemanSR17,DBLP:conf/aaai/LiuWSCS18,DBLP:conf/emnlp/GongFQL19,DBLP:journals/corr/abs-2004-14373}, AMR-to-text generation \cite{song-etal-2018-graph,damonte-cohen-2019-structural,mager-etal-2020-gpt,zhao-etal-2020-line}, graph-to-text generation \cite{yao-etal-2020-heterogeneous,zhao-etal-2020-bridging,song-etal-2020-structural}, and so on.
Though achieving good performance on surface-level NLG, they still suffer from low fidelity and uncontrollable content selection \cite{DBLP:conf/acl/ChenCSCW20}. 

To address this issue, it is intuitive to leverage external logical knowledge for better generation \cite{DBLP:conf/emnlp/ZhangDLCZC20,DBLP:journals/corr/abs-2104-07650,DBLP:conf/acl/DengZLHTCHC20,DBLP:journals/corr/abs-2109-08306,DENG2022107584}. 
\cite{DBLP:conf/acl/ChenCSCW20} firstly proposes text generation using logical inferences from a table. 
Their study mainly supports probing purposes or evaluates neural models' ability to generate logically correct descriptions based solely on the table content.
Note that the best model in~\cite{DBLP:conf/acl/ChenCSCW20} only achieves better than  \textbf{20\%}  factual correctness rate according to a follow-on human evaluation. Thus, the formulation of this approach still misses the mark for real-world text generation systems due to the low fidelity and uncontrollability.
Text2Logic \cite{DBLP:conf/emnlp/ChenCZZZSW20} formulates NLG as a logical form to the text generation problem. 
Alongside the table information, the model is provided with the logical form. 
However, its performance relies on the availability of large numbers of supervised data (i.e., logic--text pairs), thus restricting its applicability. 

Our work relates to the few-shot NLG. 
TableGPT \cite{DBLP:conf/coling/GongSFQBLL20} focuses on generating high-fidelity text for the table-to-text generation using limited training pairs. 
Another work \cite{DBLP:conf/acl/ChenECLW20}  propose a few-shot NLG approach with language modeling to compose coherent sentences with content selection.
However, those approaches are trained and tested mainly on surface-level descriptions, which are not straightforwardly applicable to the logical knowledge-conditioned text generation. 
From a methodological perspective, our work relies on self-training  \cite{DBLP:journals/kais/TrigueroGH15,DBLP:conf/nips/LiSLZZCS19} which has shown some surprising success with natural language processing (NLP) tasks \cite{DBLP:journals/corr/abs-2006-15315,DBLP:conf/emnlp/MengZHXJZH20,DBLP:journals/corr/abs-2010-02194,DBLP:conf/ijcai/QiZCCXZZ21}.
Our work also relates to dual learning \cite{DBLP:conf/nips/HeXQWYLM16,schmitt-etal-2020-unsupervised} which tackle the training data bottleneck through a dual-learning game. Differently, we integrate the dual tasks into the self-training framework.

\section{Methodology}
\subsection{Problem Definition}
The goal of logical knowledge-conditioned text generation is to generate natural language $Y$ from tables $T$ conditioned on logical forms $L$. 
{\color{highlight} Given an input table $t_i \in T$ with a logical form $l_i \in L$ as a condition, we follow \cite{DBLP:conf/emnlp/ChenCZZZSW20} to linearize the table content  $t$, and the logical form  $l$ and then concatenate them to obtain the input sequence $x_i$.
}
In the few-shot setting, we have an extremely small paralleled dataset with $T = \{x_i,y_i\}_i^{N}$ and many unlabeled texts $U = \{u_j\}_j^{M}$, where $M \gg N$. 
Note that it is easy to obtain a large scale of unlabeled and diverse text corpus, but rather difficult to acquire their corresponding logical forms.
Our target is the mapping function, $Logic2Text$, between $x_i$ and $y_i$. Formally, we have:
\begin{equation}
    y_i = Logic2Text(x_i,\phi),
\end{equation}
where $x_i$ is the concatenation of input logical form and table content (\textbf{including table captions and headers}), $\phi$ is the parameter of $Logic2Text$, and $y_i$ is the output text. 
\subsection{Framework}
As shown in Figure \ref{arc}, we regard logical knowledge-conditioned text generation  as a sequence-to-sequence task and introduce the encoder and decoder architecture in \S\ref{encoder} and self-training in \S\ref{self}. 
To select high-quality samples from pseudo-logical forms, we introduce the content consistency module, which leverages the reverse task of $Text2Logic$ to estimate the semantic consistency score ($text \rightarrow logic \rightarrow text^{\prime}$) in \S\ref{sec1}. Furthermore, we introduce a structure consistency module with rules to score those instances in \S\ref{sec2}. Finally,  we introduce the overall optimization procedure and training details in \S\ref{sec4}. 

\subsection{Encoder--Decoder for Logical Knowledge-Conditioned Text Generation}
\label{encoder}
We utilize the pre-trained language model as an encoder. Specifically, we leverage the generative pre-trained transformer GPT-2 \cite{radford2019language} as the backbone, following \cite{DBLP:conf/acl/ChenECLW20}. Note that our approach is model-agnostic, and other architectures, such as UniLM \cite{DBLP:conf/nips/00040WWLWGZH19}, and BART \cite{DBLP:conf/acl/LewisLGGMLSZ20}, can be applied. We concatenate the table content $t$ and logical forms $l$ as input sentences following \cite{DBLP:conf/emnlp/ChenCZZZSW20}. 
We leverage the transformer to encoder each sentence as vectors.
We utilize the same architecture with different parameters for the pseudo logical form generation (i.e., $Text2Logic$ generation). Because lots of output sequences share the same tokens with the nodes in the input logical trees, we introduce a logic-tree-based copy mechanism for decoding.

\subsubsection*{Logic-tree-based Copy Mechanism}

{\color{highlight}
We first leverage a gate that decouples the framework into language-model-based generation and tree-node selection~\cite{DBLP:conf/acl/SeeLM17}. 
We leverage a soft gate, $p_{copy}$, to choose between copying from \emph{logic-tree nodes} using attention weights as the probability distribution or generating from softmax-over-vocabulary:}
\begin{equation}
\begin{aligned}
    &p_{copy} = \sigma(W_cc_t + W_ss_t + W_xx_t + b),
\end{aligned}
\end{equation}
where $x_t, s_t$ are the decoder input, state, respectively. 
{\color{highlight} $\sigma$ refers to the sigmoid activation function.}
$c_t = \sum_{i}a_t^ih_i$ and $h_i$ is the encoder hidden state at time step $t$. $W_c, W_s, W_x$, and $b$ are trainable parameters. We optimize the copy probability, $p_{copy}$, using an additional loss as follows:
\begin{equation}
\begin{aligned}
    &L = L_c + \lambda\sum_{\begin{subarray}{c}w_{j} \in V_i \end{subarray}} ( 1 - p^j_{copy} ), 
\end{aligned}
\end{equation}
{\color{highlight} where $L_c$ is the cross-entropy loss  (original loss) between the model outputs and target texts,}
 $\{V_i\}$ is the input logic-tree-node list, $w_j$ is the target token at position $j$, and $\lambda$ is a hyperparameter of the weight for this copy-loss term.  

\subsection{Self-training}
\label{self}
In a vanilla self-training framework, a tagger is first initialized using a set of instances having gold labels. Then, the tagger is used to tag a set of unlabeled data, and the tagging confidence for each unlabeled instance is evaluated. The automatically labeled instances having the highest confidence is added to the training set using the labels predicted by the tagger. Correspondingly, the unlabeled instance is removed from the unlabeled dataset. The tagger is then retrained using the updated training dataset and is used to tag and select the unlabeled instances from the remaining dataset.
In this paper, we leverage the $Text2Logic$\footnote{We have experiment with $Logic2Text$ with automatically produced logical forms, but obtain little improvement.} model as the tagger because 1) the unannotated texts related to tables are diverse and easy to obtain; 2) the quality of logical forms are easier to control; 3) the simplicity of applying to existing data-to-text datasets with only a few annotated logical forms. In the next section, we introduce the quality-control strategy of content and structure consistency.

\subsubsection{Content Consistency}
\label{sec1}
Inspired by back translation \cite{DBLP:conf/acl/SennrichHB16,DBLP:conf/emnlp/LampleOCDR18,DBLP:conf/aclnmt/HoangKHC18}, we leverage the reverse task of $Text2Logic$ to estimate the semantic consistency score. Back translation is proposed for machine translations wherein the sentence in the source language (e.g., Chinese) would be translated to the target language (e.g., English) and would then be translated back to the source language (e.g., Chinese). The essence of semantic consistency is that a variable, $x$, and a bijective mapping function, $f(·)$, should satisfy $\boldsymbol{x}=f^{-1}(f(\boldsymbol{x}))$, where $f^{-1}$ is the inverse function of $f$. Formally, given the pre-trained $Text2Logic$ and $Logic2Text$ models and text sample $u \in U$, we have
\begin{equation}
\hat{x} = Text2Logic(u),
\end{equation}
\begin{equation}
u^{\prime} = Logic2Text(\hat{x}).
\end{equation}
Specifically, given the original text $u$, and the recovered text $u^{\prime}$, we obtain the semantic consistency score as: 
\begin{equation}
    score_{content} =     \frac{\left(1+\beta^{2}\right) R_{lcs} P_{lcs}}{R_{l c s}+\beta^{2} P_{lcs}},
    \label{semantic}
\end{equation}
{\color{highlight} where $R_{lcs}$ and $P_{lcs}$ refer to the longest common subsequence regarding $u$ and $u^{\prime}$, respectively. 
Specifically, we obtain $R_{lcs} =  \frac{LCS(u,u^{\prime})}{len(u^{\prime)}}$ via calculating the longest common subsequence regarding $u$ and $u^{\prime}$, and obtain $P_{lcs} =  \frac{LCS(u,u^{\prime})}{len(u)}$ via calculating the longest common subsequence regarding $u^{\prime}$ and $u$.
$\beta$ is a hyper-parameter, and we utilize a development set to tune optimized $\beta$. 
}
Because logical forms are a specific data form that differs from raw text, they should obey some structure constraints. For example, the function in the generated logical form should appear in the pre-defined schema, and the number of parameters (i.e., child nodes in the logic tree) should follow the function definition. Thus, we introduce a structure-consistency module to score these instances. 

\subsubsection{Structure Consistency}
\label{sec2}

For a generated logical form $L$, we design several general rules to estimate the logical rationality score. Note that these rules are orthogonal to different types of logical forms, such as $\lambda$-calculus and Prolog. Thus, they can be easily applied to other datasets.

\textbf{Rule1: Logic Consistency} \emph{Given the generated logical form $L$, if the parentheses do not match, then the rule does not hold.}

\textbf{Rule2: Function Mutual Exclusion}. \emph{Given the generated logical form $L$, with the entire function set $O$, and the default function set $F$ (e.g., \emph{argmax}, \emph{sum}. The default operation set is defined based on the schema), if $\exists o \in O$ and $o \notin F $, then the rule does not hold.} 

\textbf{Rule3: Parameter Consistency}. \emph{Given the generated logical form $L$, with the entire function set $O$, and the default function set $F$, if $\exists o \in O$ and $nodes_{O}(o) \neq nodes_{F}(o) $, the logical rationality score is zero. Otherwise, the score is one. $nodes_{O}$ and $nodes_{F}$ denote the number of parameters of the generated and default function $o$. We calculate the average parameter consistency for all nodes in the logic tree using the breadth-first search. If the average score is lower than $\kappa$, then the rule does not hold.}

\subsection{Training Details}
\label{sec4}
After obtaining the content consistency and structure consistency scores, we utilize an instance-sampling approach to select the top-$K$ instances. 
The overall algorithm is shown in Algorithm \ref{alg}

\begin{algorithm}[th]
\begin{algorithmic}[1]
\caption{Self-training with Content and Structure Consistency} 
\Require train set $T=\{x_i,y_i\}_i^{N}$ with text and logic pairs, text corpus $U=\{u_j\}_j^M$, $\lambda$, $K$
\Require random shuffle $U$;
\While{$U$ is not empty}
\State $Text2Logic() \longleftarrow \operatorname{train}(Y,X)$
\State $Logic2Text() \longleftarrow \operatorname{train}(X,Y)$
\State $\hat{X} \longleftarrow \operatorname{tag}(Text2Logic, U)$
\State $U^{\prime} \longleftarrow \operatorname{tag}(Logic2Text, \hat{X})$
\State Calculate $score_{content}$ with Eq. \ref{semantic}
\State Sample top $K$ Instance $\hat{u}$ with $score_{content}$ obeying \textbf{Rule 1,2, and 3} from $U$
\State $T \leftarrow T \cup \{\hat{x},\hat{u}\}$
\State $U \leftarrow U \backslash \{\hat{x},\hat{u}\}$
\EndWhile
\State Return Logic2Text()
\label{alg} 
\end{algorithmic}
\end{algorithm} 

\begin{table*}[htbp]
\begin{center}
\small
\begin{tabular}{l|cccc|cccc|cccc}
\toprule
 \# Training instances& \multicolumn{4}{c|}{20}  & \multicolumn{4}{c|}{100} & \multicolumn{4}{c}{500}\\
 \midrule
 Metrics&B-1& R-1 & R-2 & R-L  & B-1& R-1 & R-2 & R-L & B-1& R-1 & R-2 & R-L \\
\midrule
TableGPT         &14.29&16.25& 2.54&15.31&23.02&24.61& 4.33&21.58&27.52&28.23& 6.67&25.15\\
\midrule
Seq2Seq+att      &13.31&13.59& 2.39&14.54&23.87&25.13& 3.67&21.33&31.13&33.16&10.35&30.33\\
Transformer+copy &15.35&16.87& 3.56&15.87&26.98&27.35& 5.77&23.25&33.51&35.15&12.35&32.45\\
BART             &23.51&24.12&13.35&20.14&37.32&39.31&16.71&33.31&43.03&42.17&20.35&37.25\\
GPT-2            &23.75&24.13&14.91&21.19&47.33&48.16&23.60&38.54&54.89&55.56&29.82&45.60\\
\midrule
{\color{highlight} LOGEN (BART)}
&{\color{highlight}42.13}&{\color{highlight}45..02}&{\color{highlight}20.14}&{\color{highlight}40.02}&{\color{highlight}49.34}&{\color{highlight}51.30}&{\color{highlight}29.18}&{\color{highlight}49.44}&{\color{highlight}51.32}&{\color{highlight}52.89}&{\color{highlight}32.19}&{\color{highlight}51.53} \\
LOGEN (GPT-2)&47.32&49.03&24.35&43.64&56.33&57.23&31.10&51.10&57.32&59.05&34.08&53.18 \\
\bottomrule
\end{tabular}
\caption{Evaluation results (\%) on  under 20/100/500-shot setting.}
\label{table:res_full}
\end{center}
\end{table*}

\section{Experiment}
\subsection{Dataset and Metric}
We evaluate our approach on the benchmark dataset, \textsc{Logic2Text} \cite{DBLP:conf/acl/ChenCSCW20}. We employ seven types of the most commonly used logics \cite{DBLP:conf/acl/ChenCSCW20}: \emph{count, superlative, comparative, aggregation, majority, unique,} and \emph{ordinal}. 
The \textsc{Logic2Text} dataset contains 7,566, 1,000, and 1,095 samples for training, validation, and testing, respectively. The maximum length of a natural-language segment in the dataset is 130 words, and more than 90\% of the data items are less than 90 words in length. The overall statistics of the \textsc{Logic2Text} dataset are shown in Table \ref{dataset}.
For automatic evaluations, we employ BLEU-1\footnote{Standard script NIST mteval-v13a.pl.}, ROUGE-1, 2, and L (F-measure)\footnote{rouge-1.5.5.}, noted as B-1, R-1, R-2, and R-L. 
\begin{table}[htbp]
\small
\begin{center}
\resizebox{0.45\textwidth}{!}{%
\begin{tabular}{lr}
\toprule
Tables & 5,554\\
Examples & 10,753\\
Vocabulary & 14.0k\\
Avg. description length & 16.77\\
Avg. \# nodes in logical form & 9.00\\
Avg. \# function nodes in logical form & 3.27\\
Avg. length of the linearized logical form & 24.35\\
\bottomrule
\end{tabular}
}
\caption{General statistics of \textsc{Logic2Text}.}
\label{dataset}
\end{center}
\end{table}

\subsection{Setting}
We utilize GPT-2 (124M) as the representation from \cite{radford2019language}. 
{\color{highlight} We also utilize BART-base \cite{DBLP:conf/acl/LewisLGGMLSZ20} as the backbone to  further verify the effectiveness of the framework. 
}
We employ Adam \cite{DBLP:journals/corr/KingmaB14} as the optimizer, the initial learning rate $\alpha$ is set to 2e-5. $K$ is set to 1,000, $\kappa$ is set to 0.5, and the batch size is 32. We tune the hyperparameters on the development set. We train the model on eight NVIDIA V100 16GB GPU, with the patience of 4 epochs for each iteration. 
{\color{highlight} We set a maximum number of the loop (with early stopping) in the algorithm to avoid an infinite loop. }
We run each experiment five times and calculate average performance.

\subsection{Baseline}
 \textbf{TableGPT} \cite{DBLP:conf/coling/GongSFQBLL20} We employ the TableGPT which leverages table structure reconstruction and content matching \textbf{without logical forms} for few-shot table-to-text generation. 

\textbf{Seq2seq+att} We employ the seq2seq+att with the attention model following \cite{DBLP:journals/corr/BahdanauCB14}. We concatenate the table cation, header, and the logical form as the input sequence. 

\textbf{Transformer+copy} Following \cite{DBLP:conf/emnlp/ChenCZZZSW20}, we leverage the Transformer structure with the copy mechanism as a baseline. 

\textbf{BART} \cite{DBLP:conf/acl/LewisLGGMLSZ20} We employe BART-base which is a denoising autoencoder for pretraining Seq2Seq models.

\textbf{GPT-2}\footnote{\url{https://github.com/huggingface/transformers}} We leverage the generative pre-training model GPT-2 (124M) from \cite{radford2019language}. 

We evaluate our framework LOGEN in the few-shot setting with only a few training instances (20/100/500 shots via random sampling) and regard all the other data in the training set\footnote{We leverage text which are descriptions of open-domain tables from Wikipedia.} as unannotated data\footnote{In the inference stage of a real-world NLG system, the logical forms are produced automatically based on the end applications and user interests \cite{DBLP:conf/acl/ChenCSCW20}}. 

\subsection{Main Results}
From Table \ref{table:res_full}, we observe that GPT-2 achieves better performance in all few-shot settings, as also observed by \cite{DBLP:conf/emnlp/ChenCZZZSW20}. 
We also observe that TableGPT obtains poor performance and even fails to compete with Transformer+copy, which illustrates the advantage of logic guidance. 
We notice that our LOGEN  yields better performance with \textbf{23.57\%} for B-1 \textbf{22.45\%} of Rouge-L by only \textbf{20} golden labels, which demonstrates the effectiveness of our approach. 
Note that the R-L score of fully-supervised setting with GPT-2 is 53.04 (We have reproduced this score.) \cite{DBLP:conf/emnlp/ChenCZZZSW20}, our model obtain even better performance with only 500-shot instances (\textbf{6\%} of the dataset). 
{\color{highlight} We also notice that our approach with BART-base as the backbone can yield better performance than baselines, further verifying the framework's effectiveness.}
{\color{high} We also utilize BLEU-4 for evaluation and notice that our approach with 100-shot instances obtains a comparative performance of 16.95 than GPT-2 with 17.06, illustrating the limitations of LOGEN.
We think this may cause by the few-shot data, leading to unstable performance.}

 \begin{figure*}[h]
  \centering 
  \includegraphics[width=1\textwidth]{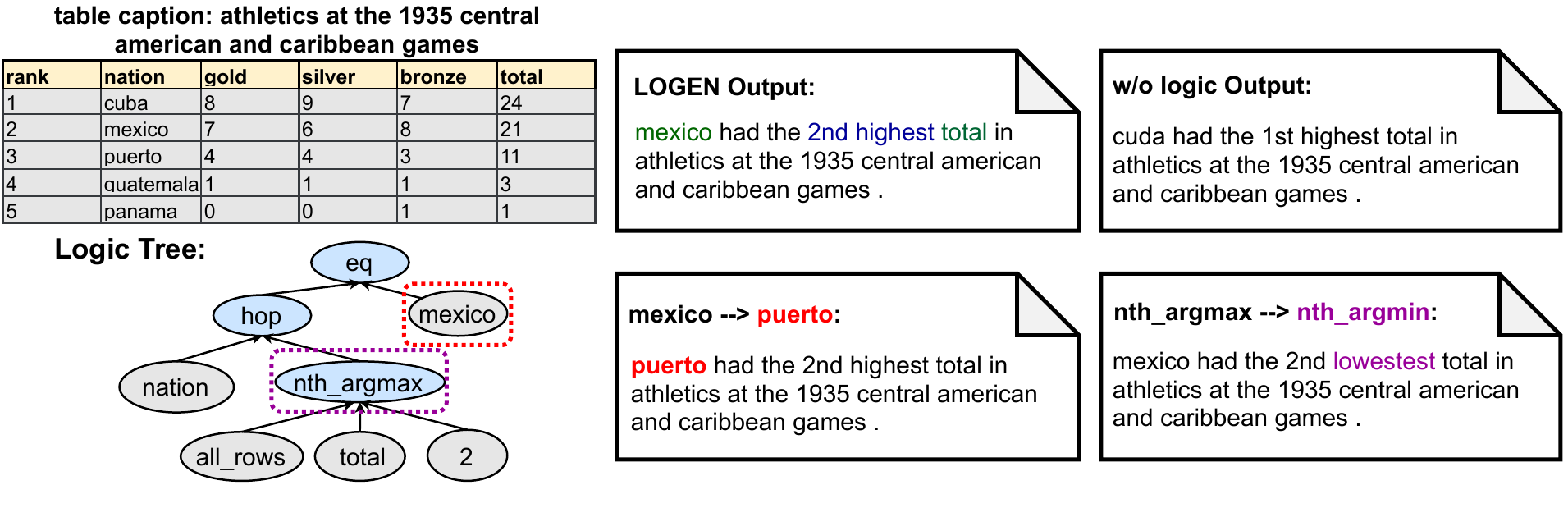} 
  \caption{Case study on our proposed LOGEN.}
  \label{case}
\end{figure*}

\subsection{Ablation Study}

\begin{figure}[h]
  \centering 
  \includegraphics[width=0.4\textwidth]{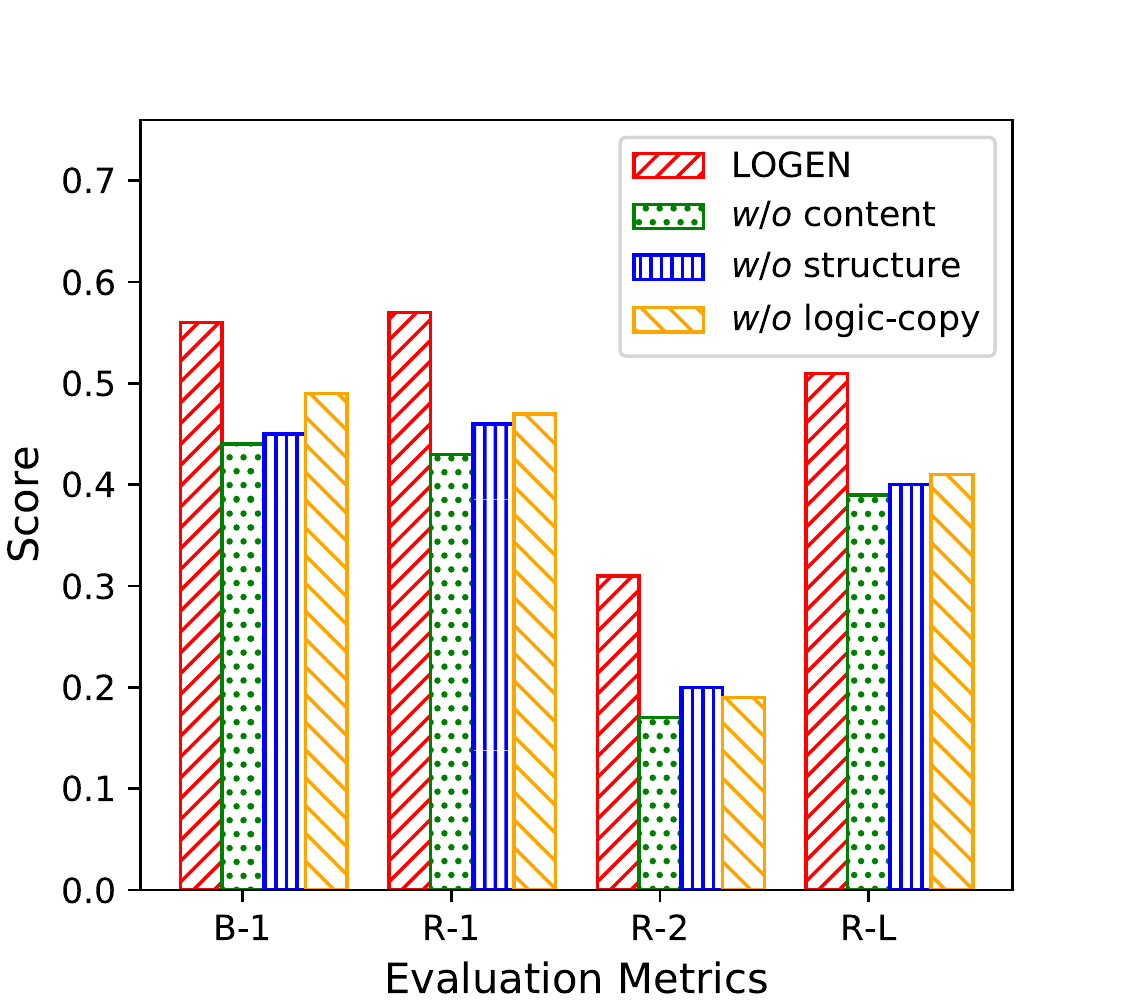} 
  \caption{Ablation study results.}
  \label{ablation}
\end{figure}

We conduct an ablation study to validate the effectiveness of the different components.
\textbf{$w/o$ content} and \textbf{$w/o$ structure} refer to a model lacking content and structure consistency, respectively.
\textbf{$w/o$ logic-copy}  refers to the model without logic-tree-based copy. 
From Figure \ref{ablation}, we observe that all models have a performance decay without content/structure consistency and logic-copy, indicating that all components are beneficial.
We also notice that the content consistency is sensitive to the Rouge score, revealing that content consistency may be more important. 

\subsection{Human Evaluation Results}

{\color{highlight}
We conduct a human evaluation to evaluate the generated answer summaries from three aspects: (1) \textbf{Informativity:} \emph{How well does the text capture the key information from the original table?} (2) \textbf{Logicalness:} \emph{How logically is the text correlated to the input logical form and table content?} and (3) \textbf{Readability:} \emph{How fluent and coherent is the text?} We randomly sample 100 instances and generate their output text using four methods (i.e., TableGPT, Seq2seq+att, GPT-2, and LOGEN) and variations of our approach ($w/o$ \textbf{all} refers to the vanilla self-training approach with GPT-2).}
{\color{high} Three data annotators with a Ph.D. degree are asked to score each generated text on a scale of 1 to 5 (higher is better). 
They are firstly trained with 100 sentences for evaluation to well understand the three metrics of informativity, logicalness and readability. 
We then ask them to annotate sampled instances to evaluate whether the there annotator could label the three metrics correctly.
We do this three times to ensure the annotator can indeed make good and consistent decisions.
We then ask the three annotators to evaluate the generated instances.
Due to the time and budget limit, we follow \cite{DBLP:conf/emnlp/ChenCZZZSW20} to sample 200 examples from each method for evaluation. 
We also calculate the average inter-rater agreement between annotators using Fleiss' kappa scores \cite{fleiss1971measuring}, finding that five of six annotations showed good agreement ($\kappa = 0.9$).}

\begin{table}[htbp]
\centering
\begin{tabular}{l|ccc}
\hline
\hline
 \textbf{Models}&\textbf{Info}&\textbf{Logic}&\textbf{Read}\\
\hline
TableGPT &2.22 & 2.20 & 3.01 \\
Seq2seq+att&2.33 &2.15 &3.13 \\
BART& 2.88&2.35 & 3.92\\
GPT-2& 2.78&2.45 & 3.88\\
\midrule
LOGEN& \textbf{3.98}& \textbf{4.54}& \textbf{4.35} \\
$w/o$ content&  3.62&  4.31&
 3.93 \\
$w/o$ structure&  3.52&  4.21&
 3.95 \\
$w/o$ all  & 3.02& 2.67&  3.90 \\
\hline
\end{tabular}
\caption{Human evaluation results.}
\label{human}
\end{table}

Table \ref{human} lists the human evaluation results, showing that our approach consistently outperforms the other methods in all aspects. 
We observe that TableGPT (\textbf{without logic}) achieves the lowest logic score. Note that TableGPT-generated text lacks an explicit logical form. Thus, the model has trouble generating logical correct text. 
Logical forms significantly affect the logic performance scores.
Seq2seq+att generate text using a sequence-to-sequence model, resulting in the low-quality text in the few-shot setting.
GPT-2 and BART achieve relatively low scores in informativity and logic, which may be caused by the failure of NLG in the few-shot setting. 
However, GPT-2 and BART generate more fluent text having higher readability scores, which may have taken advantage of the pre-trained language model. 
\textbf{$w/o$ content}  and  \textbf{$w/o$ structure} obtain a performance drop compared with LOGEN, further indicating the effectiveness of the different components.
\textbf{$w/o$ all}  obtains only a small performance improvement in the human evaluation compared with GPT-2, revealing that the quality of the self-labeled logical form influences model performance.

\subsection{Manipulating Text with Logic}

To analyze the effect of logical forms for text generation, we randomly sample from the instance and conduct an experiment\footnote{More generated examples are in the supplementary materials.}. From Figure \ref{case}, we observe that, without logic condition, the model misses some important entities or logic types (\emph{argmax}) and are logically wrong. We further notice that, when we permute the logical forms with different entities (\emph{Sep 21} to \emph{Sep 15}) or functions (\emph{argmax} to \emph{argmin}), our model generate corresponding text with the logic condition, which indicates that logical forms can guide text generation, thus, promoting the logical correctness of NLG. 

\subsection{Analysis}
\label{analysis}
\subsubsection*{Error Analysis}
\begin{figure}[!htbp]
\centering
\includegraphics[width=0.5\textwidth]{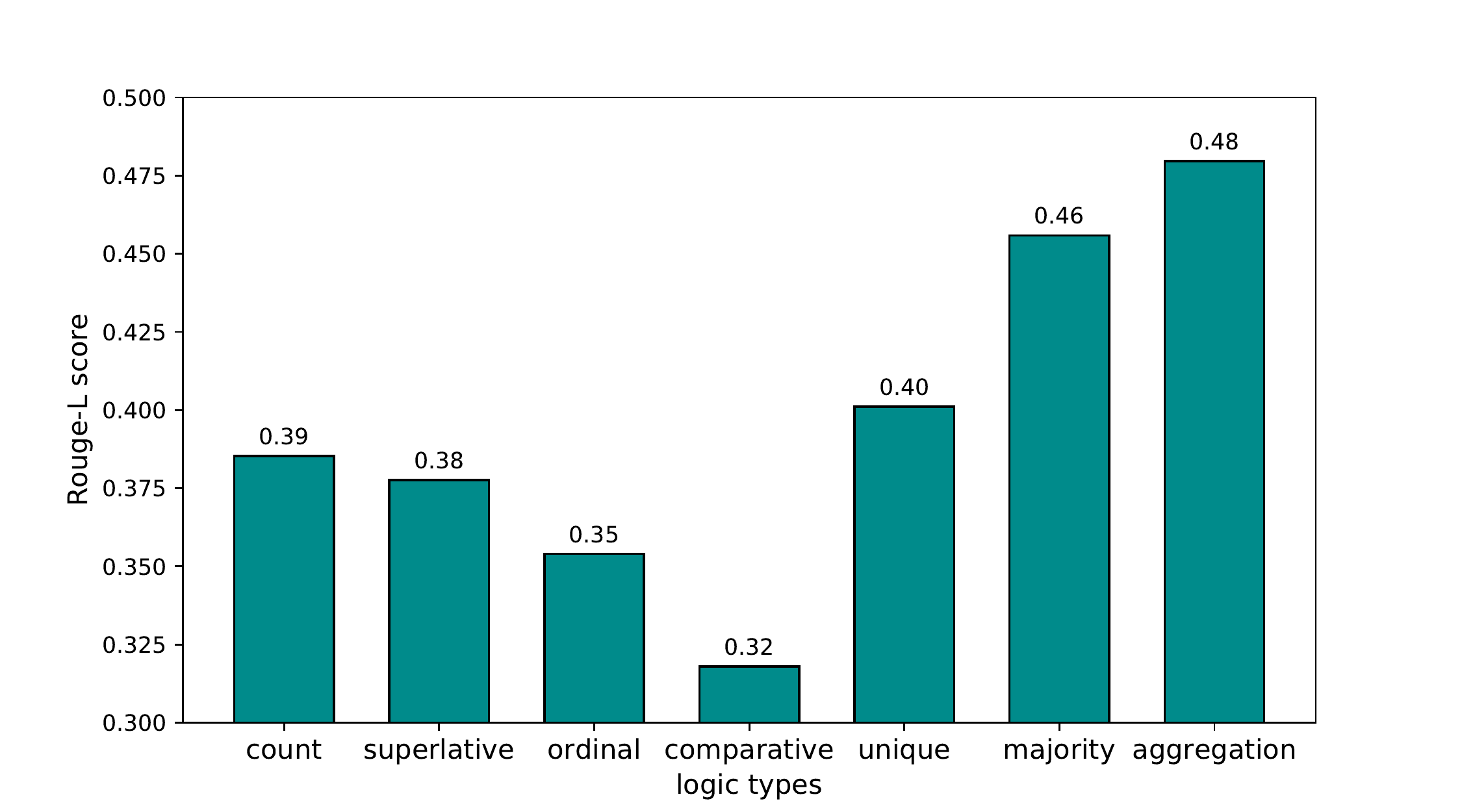}
\caption{\label{pr1} Error analysis with different logic types.}
\label{error}
\end{figure}

{\color{highlight} We conduct an error analysis of our approach. From Figure \ref{error}, we observe that text generation with the logic type of \emph{comparative} obtain the most deficient performance, indicating that the model still suffers from numerical logic reasoning. Moreover, we observe that the results of all logic types are still far from satisfactory. This indicates that few-shot generation is rather challenging and may require additional schemes and extra information to improve.}

\subsubsection*{Impact of Different $K$}
 
\begin{figure}[H]
\centering
\includegraphics[width=0.5\textwidth]{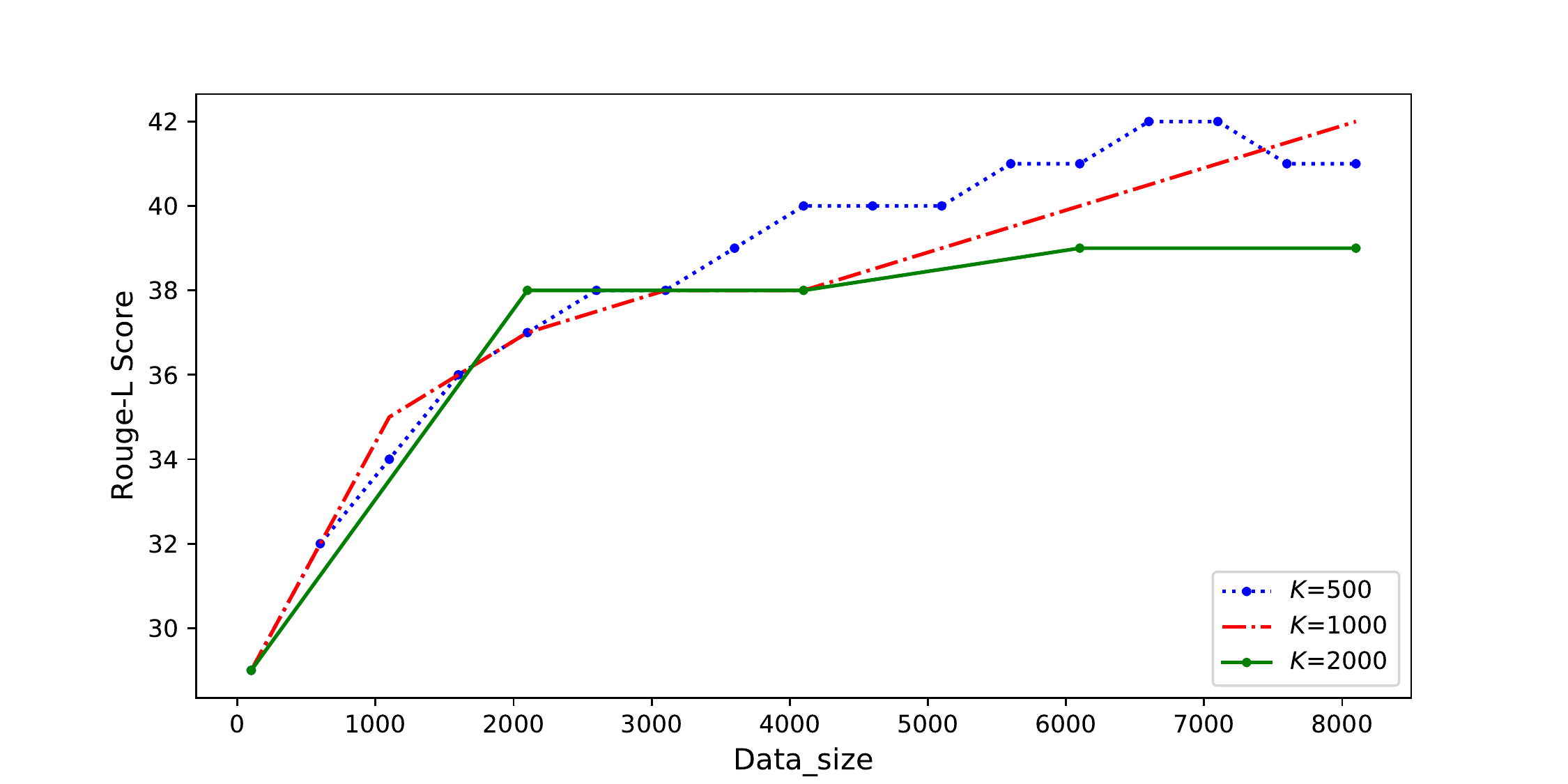}
\caption{Evaluation results with $k$, a.k.a., different sampling numbers.}
\label{number_k}
\end{figure}

Furthermore, we investigate the performance with different $K$ regarding the number of instances in each iteration. From Figure \ref{number_k}, we observe that, with an increase in the number of training steps, the model gradually achieves better performance. We also notice that the model obtain comparable performance when $K$ is 500 or 1,000. Because a small $K$ leads to more iterations that require more computing resources, we set $K=1,000$ to balance performance and computation complexity\footnote{We train the model with $K$=5,00 in more than 32 hours}. 

\subsubsection*{Impact of Different Instances }
\begin{figure}[!htbp]
\centering
\includegraphics[width=0.45\textwidth]{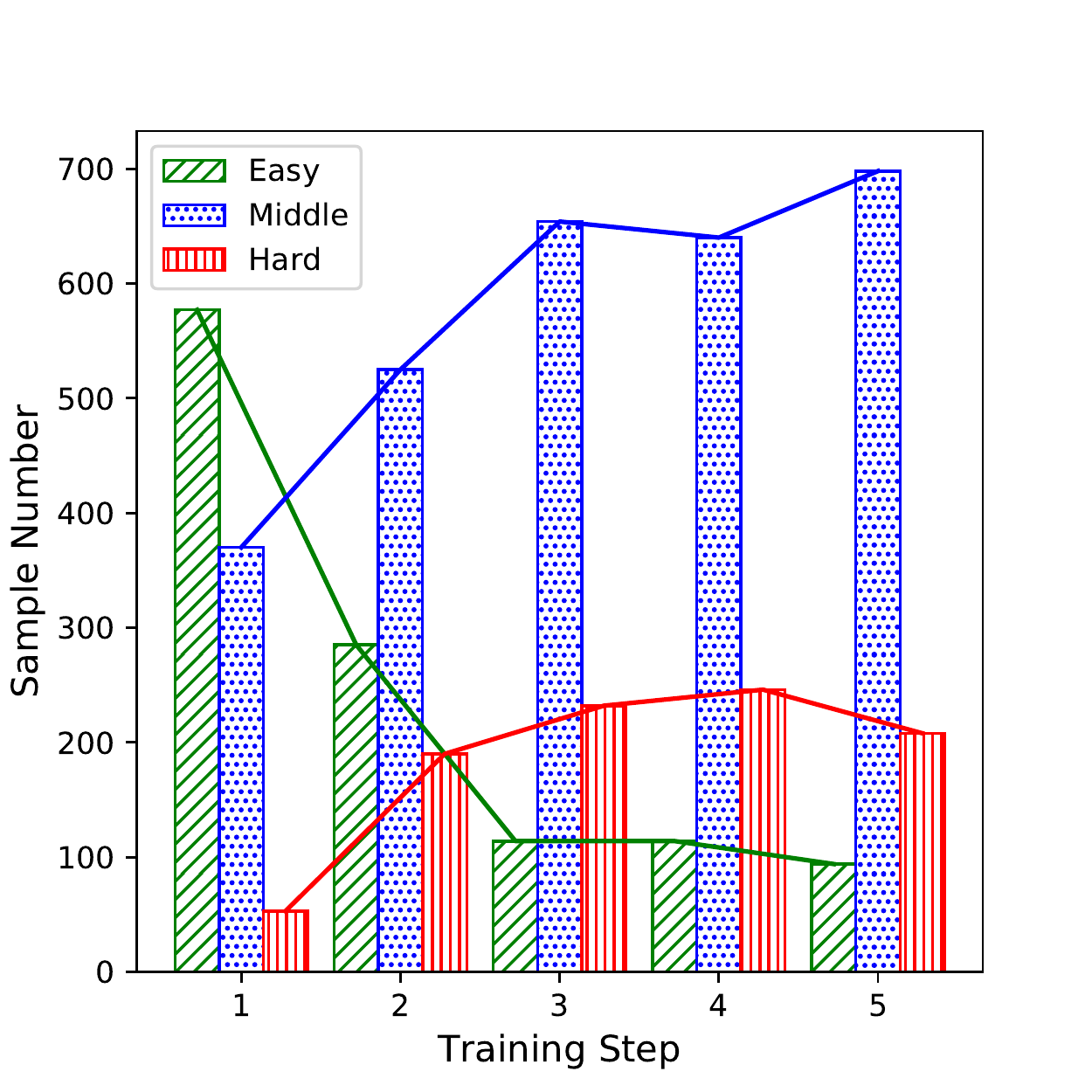}
\caption{Evaluation results of different instances in different training steps.}
\label{instance}
\end{figure}

Finally, we study the problem of which samples to choose at each iteration to promote future works. We divide the \textsc{Logic2Text} dataset into \emph{easy}, \emph{middle}, and \emph{hard} subsets based on the logical trees' layer depths. Specifically, we regard the brackets, \{ and \}, as layer dividers for the logical form. Intuitively, a sample having a large layer depth should be more complex and difficult to predict. 
{\color{highlight} We obtain 1,943 hard instances, 4,068 middle instances, and 2,555 easy instances. Figure \ref{instance} illustrates the samples chosen for each iteration by LOGEN. The green, yellow, and red bars refer to the easy, middle, and hard instances. 
We notice that, during the early stage of training, the model is prone to choosing those easy instances, whereas, during the last stage, the model mostly chooses the difficult samples. 
We think this is because, during the early stage, the model can not obtain high qualified pseudo instances with only few-shot training samples. 
When the iteration increases, the model performance increases and more hard instances can be tagged with qualified generation targets (pseudo data). 
This observation indicates that our model implicitly learns the training curriculum for self-training. }

\section{Conclusion}
This paper studies the few-shot logical knowledge-conditioned text generation problem and proposes a unified framework, LOGEN. Experimental results indicate that our approach achieves better performance than baselines on the benchmark dataset. With our approach, we successfully generate text with logic guidance using only a few seeded training instances, which can be applied to many real-world data-to-text generation applications. Our framework is general in the sense that any generation model with different logical types can be employed. In the future, we plan to study the problem of controlled NLG without logical forms (i.e., zero-shot logical NLG) and to extend our approach to more challenging tasks in which logical forms cannot be induced using a tree-style.

\section*{Broader Impact Statement}
A broad goal of NLG is to generate fully-synthetic, faithfully representative text segments to facilitate data sharing. 
For example, it is of high value in the medical domain and provides a social benefit to generate emergency department-chief complaints, a history of present illness, or the progress notes from electronic health records. 
However, previous large-scale pre-trained language model (e.g., GPT-2/3) still lack the ability to generate logical correct texts, thus, missing the mark for real-world text generation system.
Our approach can leverage only a few logical forms to generate fidelity and logically correct descriptions of these reports, promoting the fulfillment of NLG applications.
Our vision is to develop a logical controllable text generation system for the NLP community, and our innovation is a small step in that direction. 
Our framework may fail when integrated with illegal or malicious logical forms, thus, generating unintended texts. We leave this for future works.

\section*{Acknowledgments}
We want to express gratitude to the anonymous reviewers for their hard work and kind comments.
We thank Ning Ding for helpful discussions and feedback on this paper. 
This work was supported by the National Natural Science Foundation of China (No.62206246), Zhejiang Provincial Natural Science Foundation of China (No. LGG22F030011), Ningbo Natural Science Foundation (2021J190), and Yongjiang Talent Introduction Programme (2021A-156-G), CAAI-Huawei MindSpore Open Fund, and NUS-NCS Joint Laboratory (A-0008542-00-00).

\bibliographystyle{IEEEtran}
\bibliography{sample}

\vspace{-1cm}
\begin{IEEEbiography}[{\includegraphics[width=1in,height=1.25in,clip,keepaspectratio]{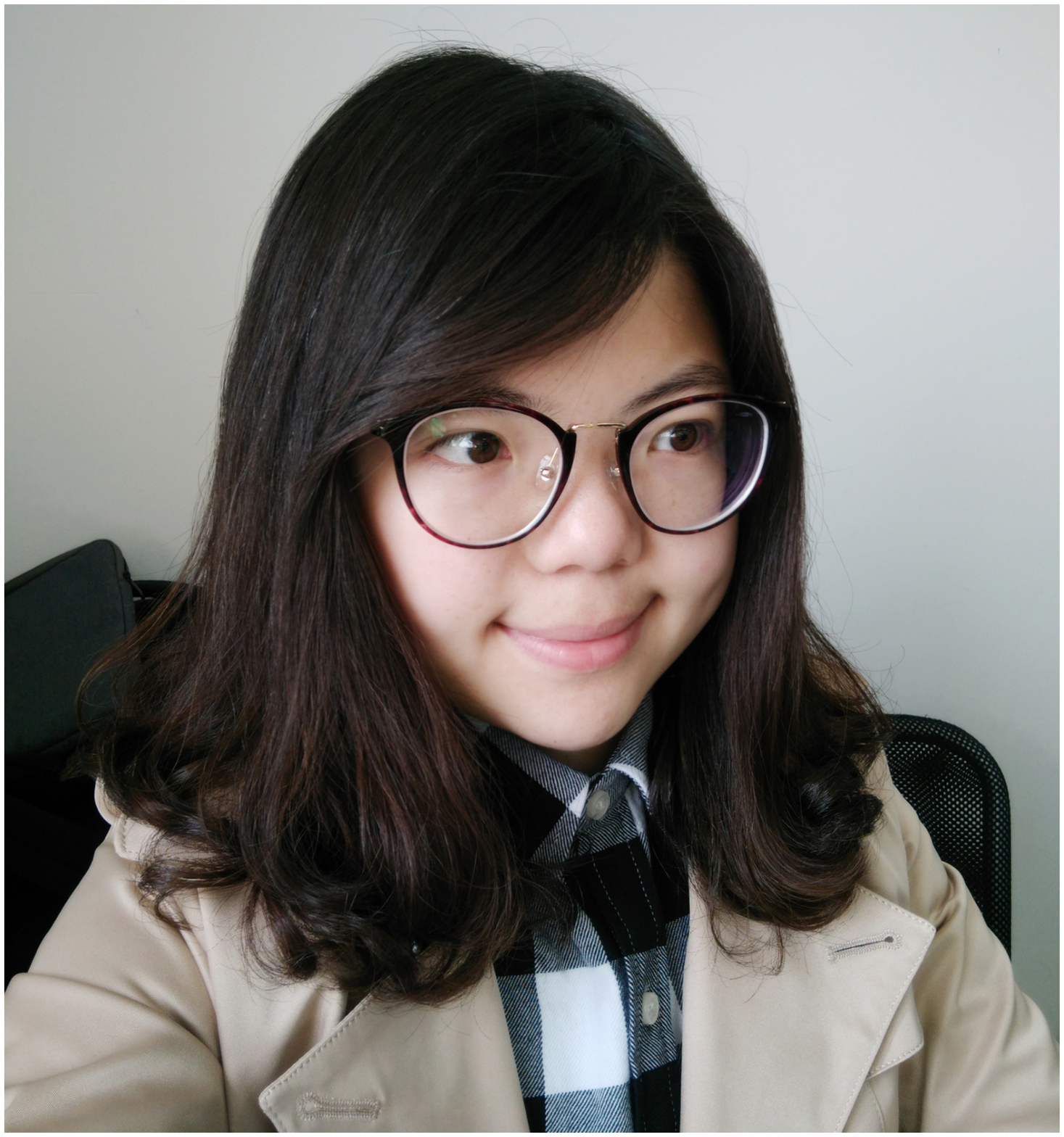}}]{Shumin Deng} is a research fellow in the Computer Science Department, School of Computing at the National University of Singapore. She obtains her Ph.D. in Computer Science and Technology at Zhejiang University (2022).
Her research relates to Information Extraction, Knowledge Graph, Knowledge Representation \& Neuro-Symbolic Reasoning. 
\end{IEEEbiography}

\vspace{-1cm}
\begin{IEEEbiography}[{\includegraphics[width=1in,height=1.25in,clip,keepaspectratio]{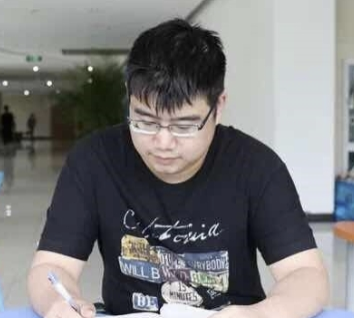}}]{Jiacheng Yang} is a master in the College of Computer Science and Technology, Zhejiang University. His research interests include natural language processing, knowledge graph.
\end{IEEEbiography}

\vspace{-1cm}
\begin{IEEEbiography}[{\includegraphics[width=1in,height=1.25in,clip,keepaspectratio]{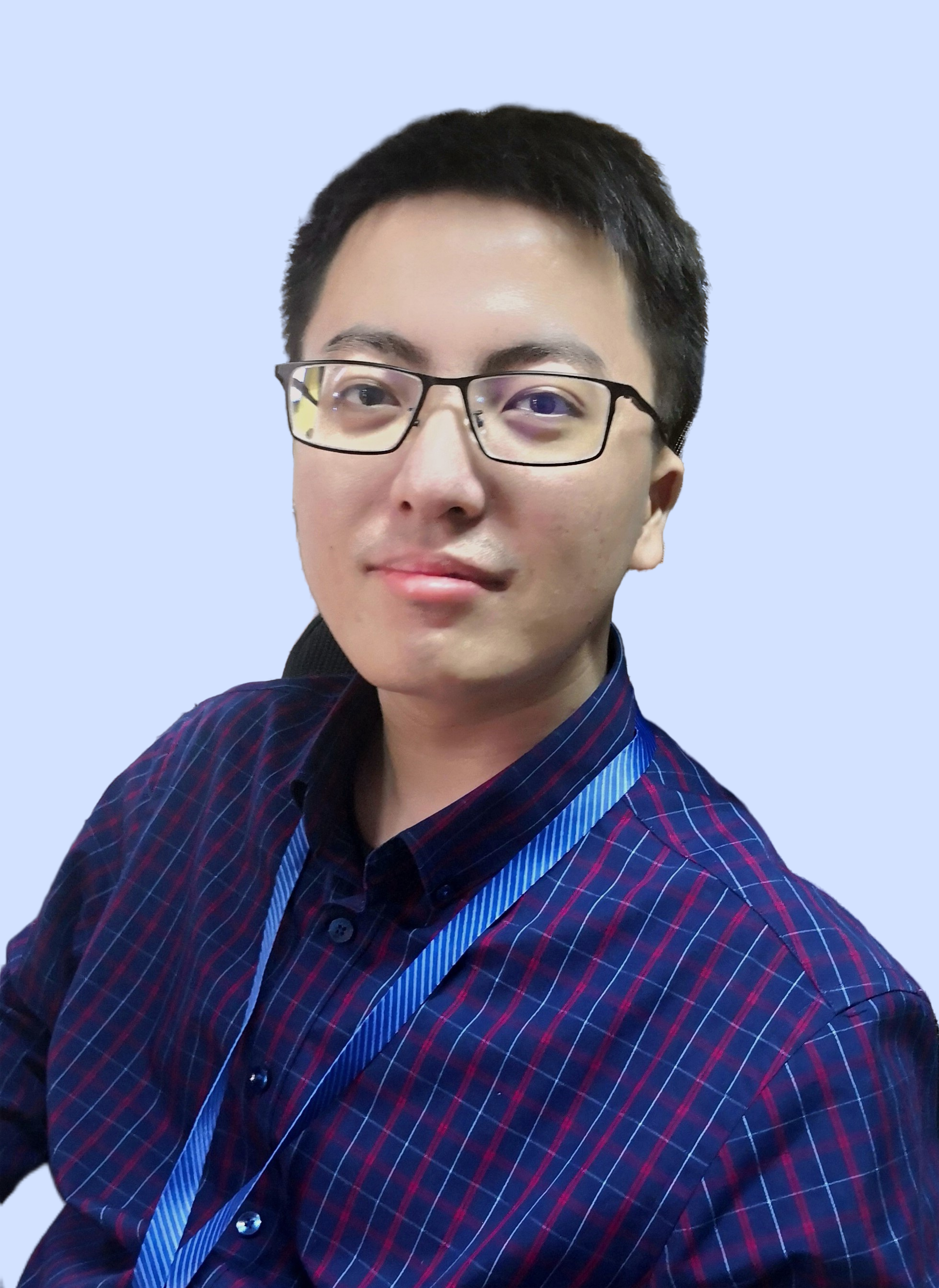}}]{Hongbin Ye} is a Ph.D. candidate in the College of Computer Science and Technology, Zhejiang University. His research interests include information extraction, knowledge graph, and controllable neural natural language generation.
\end{IEEEbiography}

\vspace{-1cm}
\begin{IEEEbiography}[{\includegraphics[width=1in,height=1.25in,clip,keepaspectratio]{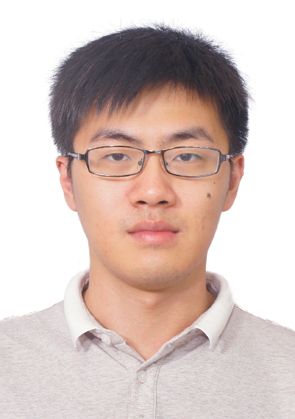}}]{Chuanqi Tan} received the Ph.D. degree in the School of Computer Science and Engineering from Beihang University, Beijing, China, in June 2019. He is currently an Algorithm Expert of Language Technology Lab, Alibaba DAMO Academy. His research interests include question answering, information extraction, and biomedical natural language processing.
\end{IEEEbiography}

\vspace{-1cm}
\begin{IEEEbiography}[{\includegraphics[width=1in,height=1.25in,clip,keepaspectratio]{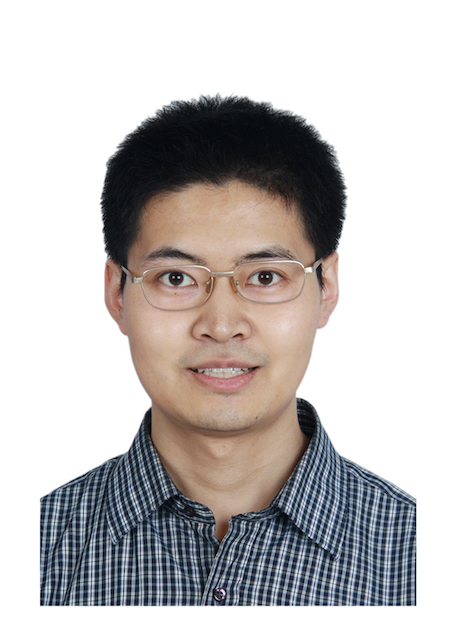}}]{Mosha Chen} received his master's degree from the Department of Computer Science and Engineering, Shanghai Jiaotong University. He is a staff algorithm engineer in Alibaba Group. His research interests include natural language processing and medical AI technology. 
\end{IEEEbiography}

\vspace{-1cm}
\begin{IEEEbiography}[{\includegraphics[width=1in,height=1.25in,clip,keepaspectratio]{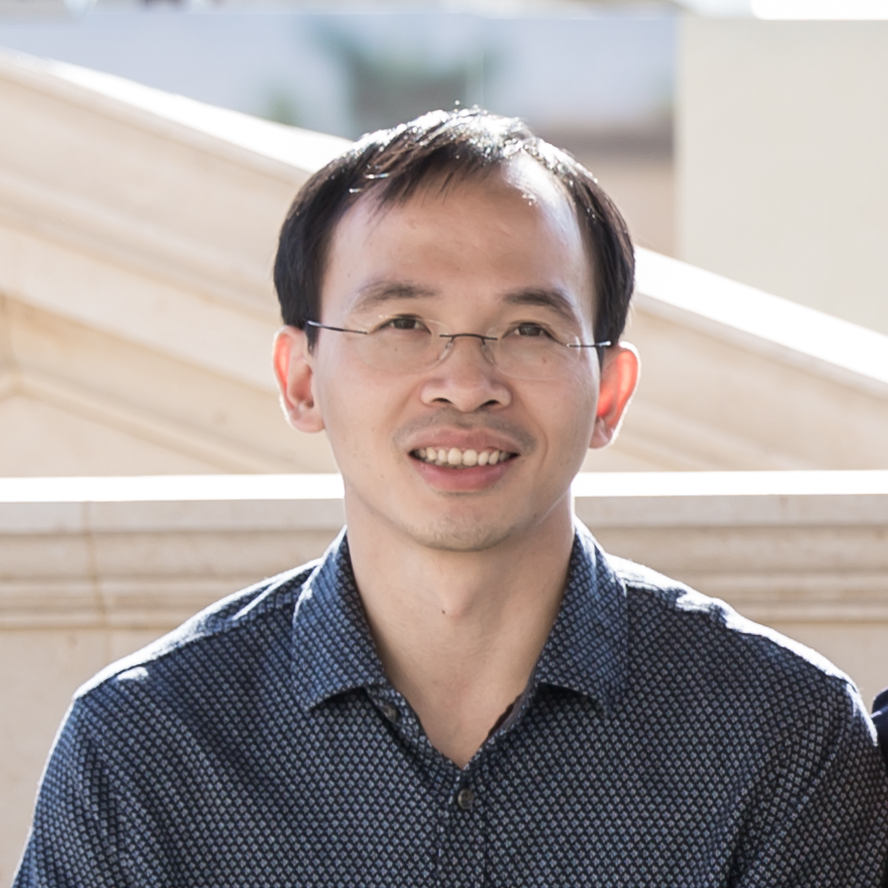}}]{Songfang Huang} is currently a Senior Staff Algorithm Engineer of Language Technologies Lab, Alibaba DAMO Academy. He leads a team working on large-scale pre-trained language models and AI for Healthcare. Before joining Alibaba, he is a research manager and research staff member of IBM Research. His research interests include language modeling, question answering, information extraction, and cognitive healthcare.
\end{IEEEbiography}

\vspace{-1cm}
\begin{IEEEbiography}[{\includegraphics[width=1in,height=1.25in,clip,keepaspectratio]{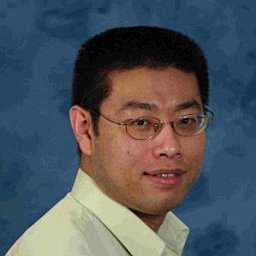}}]{Fei Huang} is a principal researcher of Language Technologies Lab, Alibaba DAMO Academy. He leads R\&D on NLP foundational technologies, dialogue, and machine translation. His team develops various NLP technologies ranging from lexical, syntactical, semantic, discourse, as well as deep learning-based algorithms, and integrate them into the Alibaba NLP platform, which supports several hundred internal and external clients with advanced NLP models, systems, and solutions in various industries.
\end{IEEEbiography}

\vspace{-1cm}
\begin{IEEEbiography}[{\includegraphics[width=1in,height=1.25in,clip,keepaspectratio]{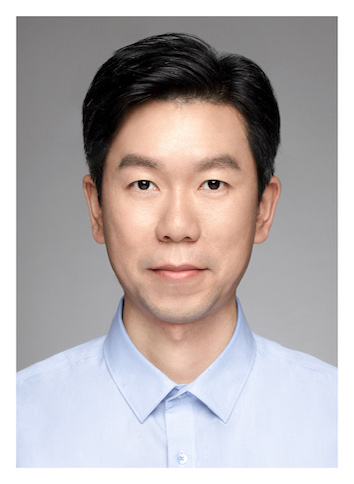}}]{Huajun Chen} is a full professor of College of Computer Science and Technologies at Zhejiang University, serve as the Director of Joint Lab on Knowledge Engine of AZFT (Alibaba-Zhejiang University Joint Research Institute of Frontier Technologies), and a deputy director of the Key Lab of Big Data Intelligence at Zhejiang Province.  He received his bachelor's degree and a Ph.D. from Zhejiang University in 2000 and 2004, respectively. He worked as a visiting assistant professor at Yale Center for Medical Informatics, Yale University (From June 2006 to June 2007), and a visiting scholar at the School of Computer Science of Carnegie Mellon University (From June 2007 to August 2008).
\end{IEEEbiography}
 
\vspace{-1cm}
\begin{IEEEbiography}[{\includegraphics[width=1in,height=1.25in,clip,keepaspectratio]{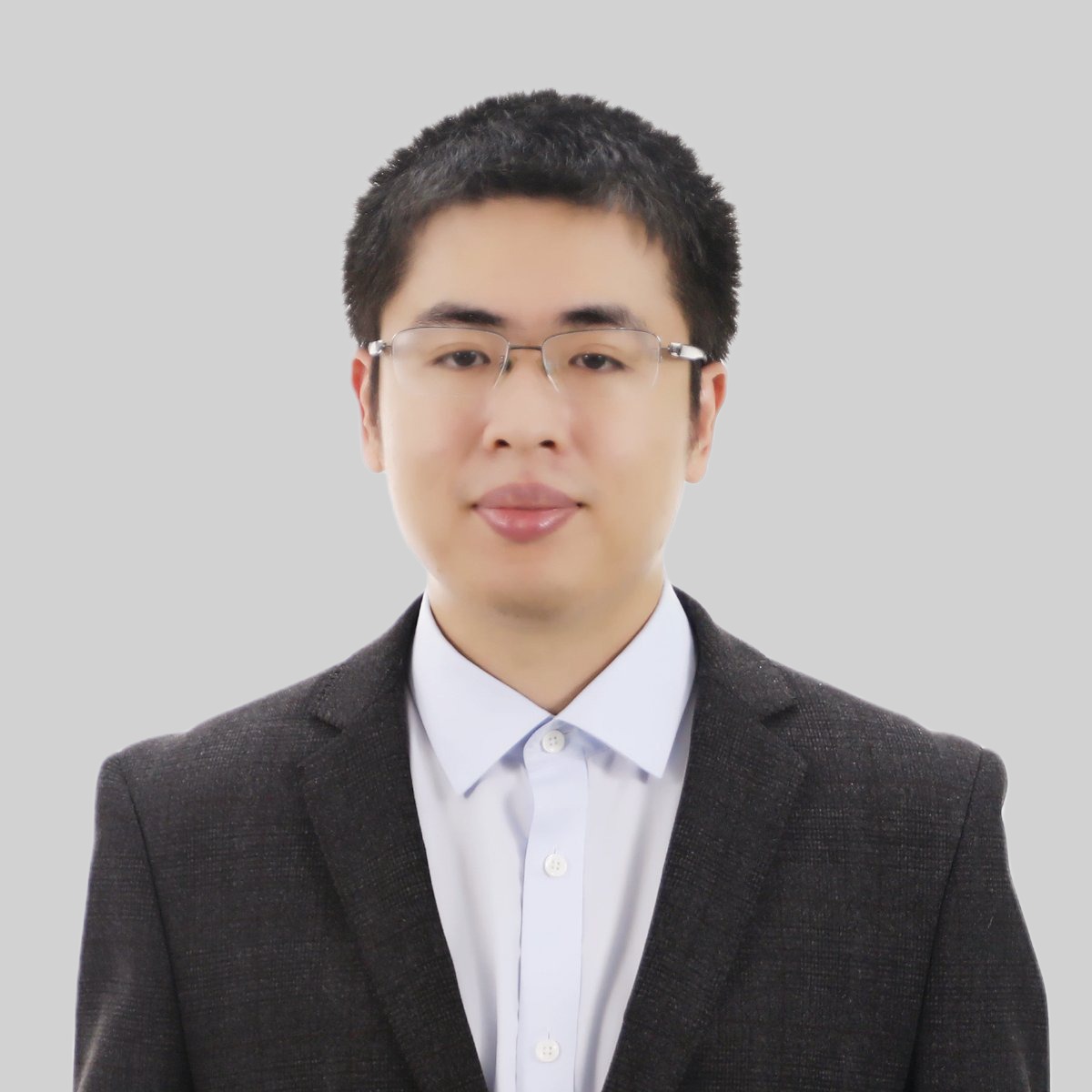}}]{Ningyu Zhang} is an associate professor/doctoral supervisor at Zhejiang University, leading the group about KG and NLP technologies.
His research interests include natural language processing, information extraction, and large language models.
He has published many papers in top international academic conferences and journals such as Natural Machine Intelligence, Nature Communications, NeurIPS, ICLR, AAAI, IJCAI, WWW, KDD, SIGIR, ACL, ENNLP, NAACL, and IEEE/ACM Transactions on Audio Speech and Language. 
He has served as Area Chair for ACL 2023, ARR Action Editor, Senior Program Committee member for IJCAI 2023, Program Committee member for AAAI, NeurIPS, ICLR, WWW, SIGIR, KDD, ICML, AAAI, and reviewer for TKDE, TKDD. 
\end{IEEEbiography}

\end{document}